\documentclass[letterpaper, 10 pt, conference]{ieeeconf}

\usepackage{graphicx}
\IEEEoverridecommandlockouts    
\usepackage{multirow}
\usepackage{algorithm}
\usepackage{algorithmic}
\usepackage{amsmath}
\usepackage{amsfonts}
\usepackage{amssymb}
\usepackage{cite}
\usepackage{url}
\usepackage{color}
\usepackage{tablefootnote}
\usepackage{tikz}

\newcommand{\ph}[1]{{\textbf{#1:}}}

\title{\LARGE \bf LOCUS: A Multi-Sensor Lidar-Centric Solution for High-Precision Odometry and 3D Mapping in Real-Time}

\author{Matteo Palieri$^{1,2}$, Benjamin Morrell$^{1}$, Abhishek Thakur$^{3}$, Kamak Ebadi$^{1}$, Jeremy Nash$^{1}$, Arghya Chatterjee$^{4}$,\\Christoforos Kanellakis$^{5}$, Luca Carlone$^{6}$, Cataldo Guaragnella$^{2}$, Ali-akbar Agha-mohammadi$^{1}$
\thanks{This work was supported by the Jet Propulsion Laboratory - California Institute of Technology, under a contract with the National Aeronautics and Space Administration (80NM0018D0004). This work was partially funded by the Defense Advanced Research Projects Agency (DARPA). \textcopyright  2020 All rights reserved.}
\thanks{$^{1}$Palieri, Morrell, Ebadi, Nash, and Agha-mohammadi are with NASA Jet Propulsion Laboratory, California Institute of Technology, Pasadena, CA, USA {\tt\footnotesize matteo.palieri@poliba.it}}
\thanks{$^{2}$Palieri and Guaragnella are with the Department of Electrical And Information Engineering, Polytechnic University of Bari, IT {\tt\footnotesize matteo.palieri@poliba.it}}
\thanks{$^{3}$Thakur is with Aptiv, Troy, MI, USA {\tt\footnotesize abhi.dtu11@gmail.com}}
\thanks{$^{4}$Chatterjee is with Bangladesh University of Engineering and Technology, Dhaka, Bangladesh. {\tt\footnotesize arghyame20buet@gmail.com}
}
\thanks{$^{5}$Kanellakis is with Luleå University of Technology, Luleå, Sweden. {\tt\footnotesize christoforos.kanellakis@ltu.se} 
}
\thanks{$^{6}$Carlone is with Department of Aeronautics and Astronautics, Massachusetts Institute of Technology, Cambridge, MA, USA. {\tt\footnotesize lcarlone@mit.edu}}
}

\begin{document}

\thispagestyle{plain}
\pagestyle{plain}

\maketitle

\begin{tikzpicture}[overlay, remember picture]
\path (current page.north east) ++(-6.3,-0.0) node[below left] {
Accepted for publication at RA-L, please cite as follows:
};
\end{tikzpicture}
\begin{tikzpicture}[overlay, remember picture]
\path (current page.north east) ++(-1.2,-0.4) node[below left] {
M. Palieri, B. Morrell, A Thakur, K. Ebadi, J. Nash, A. Chatterjee, C. Kanellakis, L. Carlone, C. Guaragnella, A. Agha-mohammadi
};
\end{tikzpicture}
\begin{tikzpicture}[overlay, remember picture]
\path (current page.north east) ++(-2.6,-0.8) node[below left] {
``LOCUS: A Multi-Sensor Lidar-Centric Solution for High-Precision Odometry and 3D Mapping in Real-Time'',
};
\end{tikzpicture}
\begin{tikzpicture}[overlay, remember picture]
\path (current page.north east) ++(-7.5,-1.2) node[below left] {
 IEEE Robotics and Automation Letters, 2020.
};
\end{tikzpicture}

\markboth{IEEE Robotics and Automation Letters. Preprint Version. Accepted November, 2020}
{Palieri \MakeLowercase{\textit{et al.}}: LOCUS: Multi-sensor Lidar Odometry and 3D Mapping}


\begin{abstract}

A reliable odometry source is a prerequisite to enable complex autonomy behaviour in next-generation robots operating in extreme environments. In this work, we present a high-precision lidar odometry system to achieve robust and real-time operation under challenging perceptual conditions. LOCUS (Lidar Odometry for Consistent operation in Uncertain Settings), provides an accurate multi-stage scan matching unit equipped with an health-aware sensor integration module for seamless fusion of additional sensing modalities. We evaluate the performance of the proposed system against state-of-the-art techniques in perceptually challenging environments, and demonstrate 
top-class localization accuracy along with substantial improvements in robustness to sensor failures. We then demonstrate real-time performance of LOCUS on various types of robotic mobility platforms involved in the autonomous exploration of the Satsop power plant in Elma, WA where the proposed system was a key element of the CoSTAR team's solution that won first place in the Urban Circuit of the DARPA Subterranean Challenge. 
\end{abstract}



\section{Introduction}

Robotic systems are rapidly entering in all aspects of human life. In particular, robots are being deployed in increasingly complex environments for a broad spectrum of applications ranging from mining~\cite{losch2018design} and search-and-rescue~\cite{jennings1997cooperative}, to industrial monitoring~\cite{bogue2011robots} and planetary exploration~\cite{haruyama2012lunar}. In these scenarios, darkness, presence of obscurants (e.g. fog, dust, smoke), lack of prominent perceptual features in self-similar areas, and slippery terrains (leading to jerky sensory motion) are common features that pose severe perceptual challenges to robotic operation. In this work, we focus on developing an accurate and reliable odometry estimation method (i.e., estimating the robot movement) which is a key requirement to enable advanced autonomous behaviours in such perceptually-challenging conditions. 

\begin{figure}[t!]
\centering
	\includegraphics[width=0.85\columnwidth, trim= 0mm 0mm 0mm 0mm, clip]{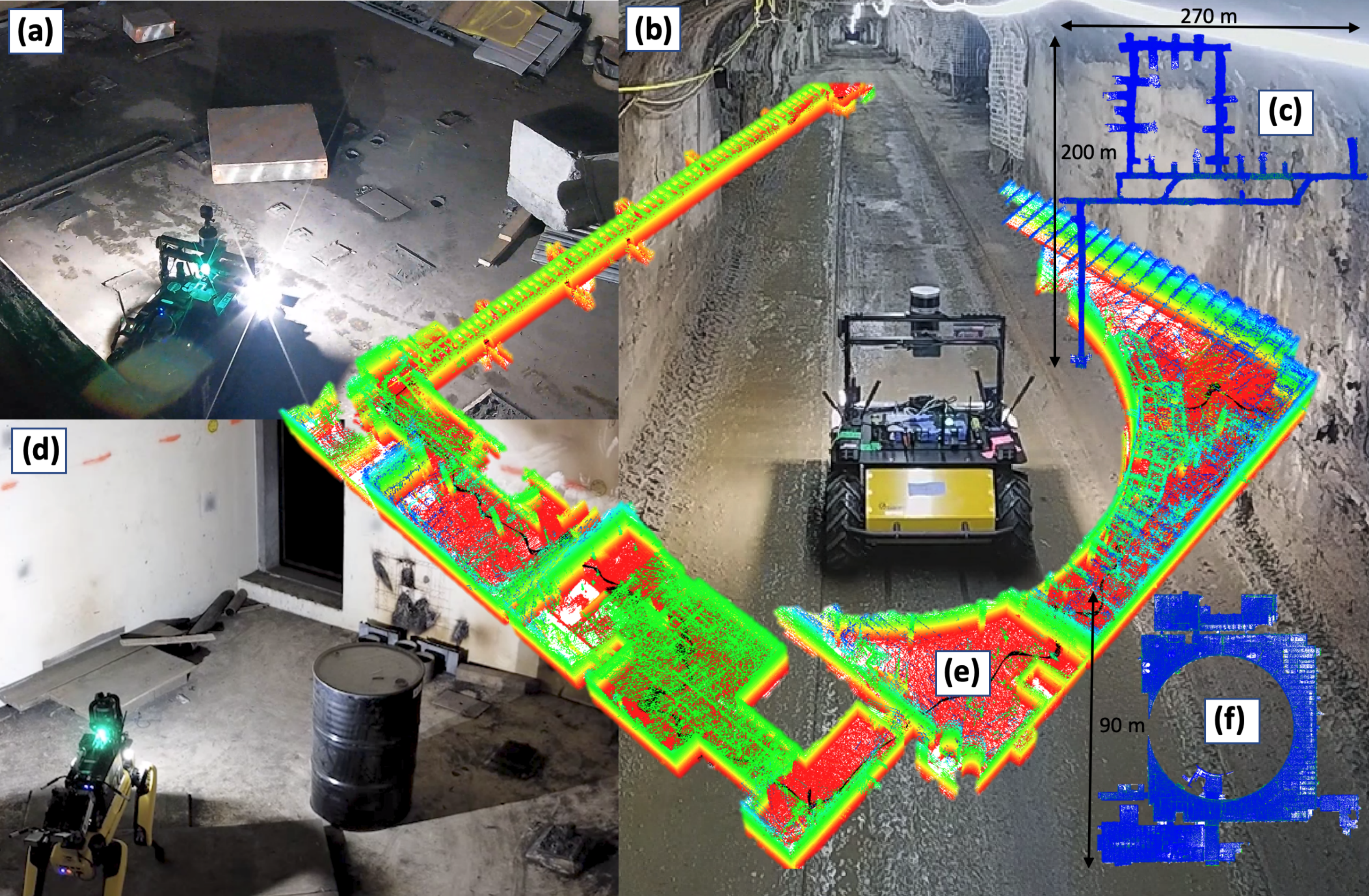} 
	\caption{Testing of the proposed lidar odometry system, LOCUS, in the DARPA Subterranean Challenge. a) Urban Circuit test environment with a Husky wheeled robot,  showing unavoidable rubble. b)  Tunnel Circuit test environment with a Husky wheeled robot carrying two lidars, showing self-similar areas with low lidar observability. c) Tunnel dataset ground truth map. d) Example Urban Circuit test with a legged robot carrying one lidar. Both robots run LOCUS. e) Map produced by LOCUS on one Husky robot in the Urban Beta course. f) Urban Alpha ground truth map.
	}
	\label{fig:cover} \vspace{-0.4cm}
\end{figure}

As lidar sensors are less sensitive to illumination variations and provide high-fidelity, long-range 3D measurements, they have been commonly used for robotic odometry estimation in the last decade. Typically, lidar odometry (LO) algorithms estimate the ego motion of the robot by comparing and registering consecutive lidar acquisitions. When it comes to perceptually-challenging settings, lidars are commonly fused with additional sensing modalities to improve accuracy. However, in these settings, potential failures in different sensing modalities can degrade, or drastically compromise the odometry performance. 

In this paper, we present LOCUS (Lidar Odometry for Consistent operation in Uncertain Settings), a lidar odometry system that \textit{(i)} enables accurate real-time operation in extreme and perceptually-challenging scenarios, and \textit{(ii)} is robust to intermittent and faulty sensor measurements. LOCUS has been a key element of the CoSTAR team's solution~\cite{CoSTAR} that won first place at the Urban Circuit of the DARPA Subterranean Challenge (SubT Challenge), where robots are tasked to autonomously explore complex GPS-denied underground environments (e.g. Fig.~\ref{fig:cover}).


\subsection{Related Works}

LO algorithms can be categorized by the representation type and the number of points (or features) used to align lidar-scans, including \textit{(i)} feature-based methods, \textit{(ii)} grid-based methods, and \textit{(iii)} dense methods.

\ph{Feature-based methods} 
Feature-based methods rely on extracting and matching salient features across consecutive lidar-scans to estimate the ego motion of the robot. Possible features can include planar and edge features~\cite{LOAM,VLOAM,MOLAblanco2019modular,le2020in2laama}, ellipsoidal surfels~\cite{Bosse} and ground features~\cite{LeGO-LOAM}. These features can be matched by proximity~\cite{LAMP}, type~\cite{LOAM}, or descriptor~\cite{Bosse}, depending on the algorithm and feature type.

\ph{Grid-based methods}
Probability grid methods map lidar-scans into grids and compute the occupancy probability densities that can later be matched using Newton's method~\cite{biber2003normal}. 

\ph{Dense methods}
Dense methods work with a large subset of lidar-scan points. As using the full point cloud can be computationally expensive for real-time operation, most approaches select a subset of points for scan matching. The Generalized Iterative Closest Point (GICP)~\cite{segal2009generalized} is a common dense point-based scan matching method, where local surface normal information is used to better address the measurement noise in scan matching by using both point-to-point and point-to-plane matching, with planes being evaluated in local neighborhoods. LOCUS falls into this category.

\ph{Scan-to-scan alignment}
The computation of the optimal alignment between two scans can be cast as a non-linear optimization problem, addressed by various solvers including Levenberg-Marquardt (e.g.~\cite{LOAM}), iterative gradient descent (e.g.~\cite{LAMP,segal2009generalized,MOLAblanco2019modular}), optimization tools such as Ceres~\cite{Ceres}, least-squares solvers (e.g.~\cite{grisetti2020solver}), or in a sliding-window, pose-graph structure (e.g.~\cite{VIL_SLAM,liosam2020shan}) with GTSAM~\cite{dellaert2012factor}.

\ph{Scan-to-map alignment}
To enable global consistency across the history of scans, the computed pose is refined by aligning the current scan and the existing map. For various map representations, ranging from feature-based (e.g. \cite{LOAM,LIO_Mapping,VIL_SLAM,LeGO-LOAM}), to grid-based maps (e.g. \cite{Cartographer}) and point-based maps (e.g. \cite{LAMP}), one can adopt different scan-to-map alignment methods. This includes point-based alignment methods (e.g. \cite{LAMP}), Normalized Distributions Transform (e.g.~\cite{biber2003normal,takeuchi20063}) or smoothing function alignment (e.g.~\cite{Cartographer}). 

\ph{Sensor fusion}
While pure lidar-based methods are powerful, their performance can significantly degrade when it comes to perceptually-challenging conditions, including environments with geometrically self-similar patterns or agile robots with high-rate motions.  
To address these challenges, it is important to fuse lidar with additional sensing modalities, such as an inertial measurement unit (IMU) or visual-inertial odometry (VIO). The IMU can provide rotational estimates that are tightly integrated~(e.g. \cite{Cartographer,LIO_Mapping,liosam2020shan,le2020in2laama}) or loosely integrated~(e.g. \cite{Bosse,VIL_SLAM,LOAM,MOLAblanco2019modular})  with the scan matching process. When the drift is translational (referred to as \textit{lidar-slip} in this paper), VIO, wheel-inertial odometry (WIO) and kinematic-inertial odometry (KIO) can complement IMUs by providing a full 6-DOF transform estimate.
Tight integration (e.g. \cite{VLOAM}) and loose integration (e.g. \cite{VIL_SLAM,Cartographer}) of these methods with lidars have shown significant improvements over individual use of either one of these modalities. LOCUS follows the loosely-coupled model, where in addition to improving accuracy, these sensor fusion methods can reduce computation by providing a near-optimal prior to the lidar scan-matching optimization in the GICP.


\subsection{Method Highlights and Contributions}

\begin{figure}[t]
  \centering
  \includegraphics[width=0.8\columnwidth, trim= 0mm 0mm 0mm 0mm, clip]{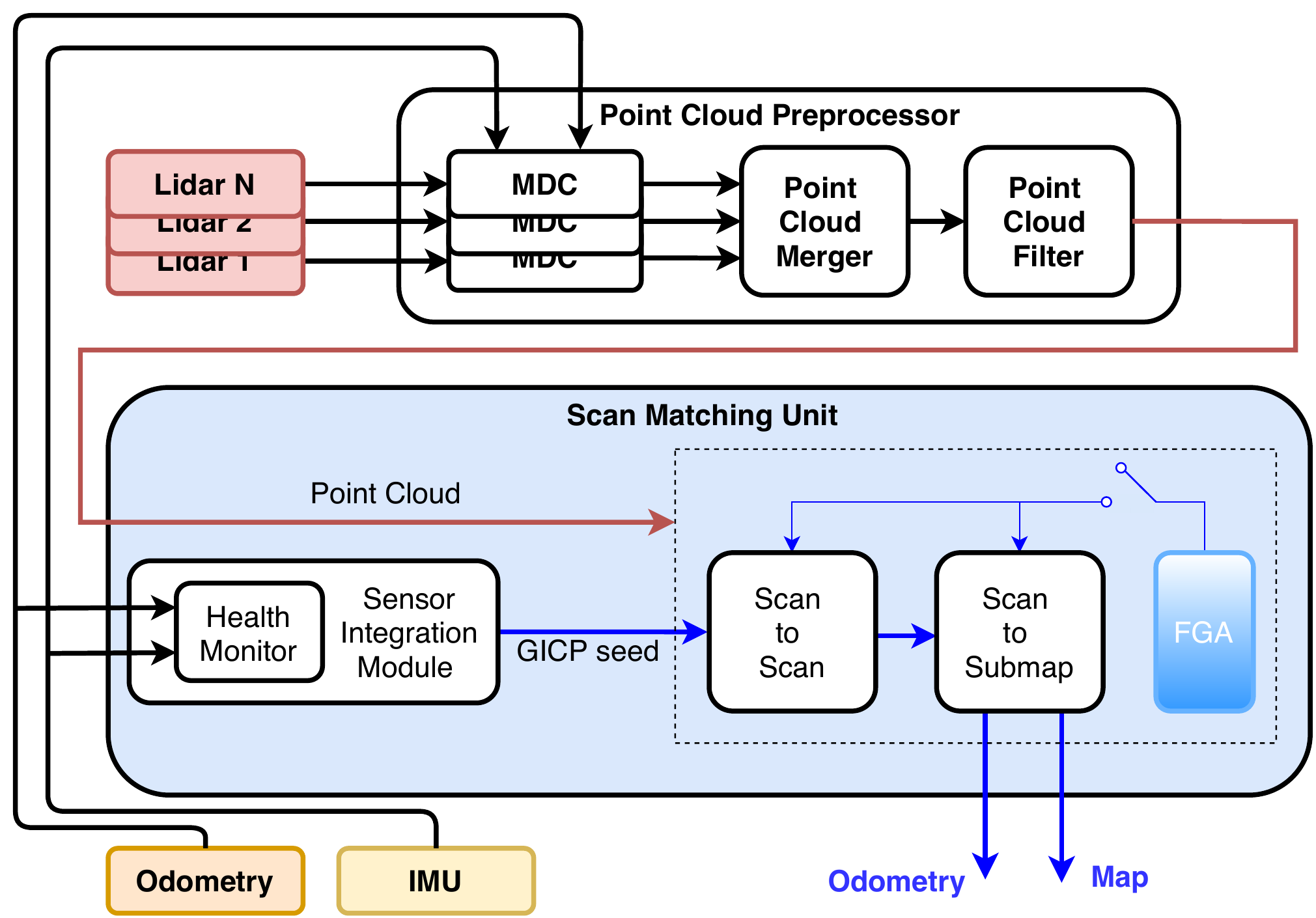}
  \caption{Architecture of the proposed lidar odometry system.}
  \label{fig:architecture}
  \vspace{-0.5cm}
\end{figure}

The highlights and contributions of LOCUS are:

\ph{1) Architecture} The LOCUS architecture (see Fig.~\ref{fig:architecture}) enables accurate, robust, and real-time odometry in perceptually-stressing settings and alleviates sensor failure challenges. The architecture can be adapted to heterogeneous robotic platforms with diverse sensor inputs and computational capabilities.
   
\ph{2) Resilience} The system is fail-safe to drops or loss of one or more sensor channels by relying on a loosely-coupled switching scheme between sensing modalities.
    
\ph{3) Environment adaptability} The system further enables incorporation of domain knowledge (if available), such as flat grounds in human-made structures. 

\ph{4) Field demonstration} We present an extensive field demonstration of LOCUS. In particular, we provide results and insights from deploying LOCUS as part of the CoSTAR team's solution that won the Urban Circuit of the SubT Challenge. We present an ablation study on LOCUS and then compare the performance with six state-of-the-art methods using the data acquired in the field tests.     

\section{System Description}\label{sec:locus}

In this section, we describe the system architecture reported in Fig.~\ref{fig:architecture} and provide details of each submodule.

\subsection{Point Cloud Preprocessor}
\ph{Motion Distortion Correction}
We assume information from one or more 360 degree lidar sensors, such as the Velodyne Puck or Ouster lidars. The raw information coming from the lidar is fed into a motion distortion correction (MDC) unit which corrects the Cartesian position of each point to account for the motion of the robot while a single scan of the lidar is completed\footnote{The Velodyne Puck lidar requires 0.1 s to complete one scan}. This correction is particularly important for points at large range, when high-rate rotations are experienced by the robot, and is a commonly employed step~\cite{LIO_Mapping,VIL_SLAM}. The correction is informed by either an IMU, or an odometry source (e.g VIO, WIO, KIO) where the chosen source depends on what is reliably available and calibrated on a given robot. 

\ph{Point Cloud Merger}
For robots with multiple lidars, to enlarge the overall robot field-of-view, the point cloud merger (Fig.~\ref{fig:architecture}, top and middle) combines each motion-corrected point cloud into a single one, using the known rigid body transformation between sensors. This step is implemented so the rest of the pipeline is consistent for robots with one or multiple lidar sensors. 

\ph{Point Cloud Filter}
The resulting point cloud is then processed in the point cloud filter to remove noise and out-of-range points,   
manage the volume of data, and reduce computational load. The point cloud filter is composed by a sequential combination of a 3D voxel grid filter and a random downsampling filter, which can be individually tuned, activated and deactivated. The voxel grid filter takes the average of the points in each 3D volume (a voxel) to decrease the data size while still capturing the dominating structure of the environment. We use a voxel size of 0.1m in the tests presented in this paper. For the random downsampling filter, we use an implementation of~\cite{vitter1984faster} with a downsampling percentage of 90\%. For both filters we use the implementation in the Point Cloud Library~\cite{rusu20113d}.

\subsection{Scan Matching Unit} 
The scan matching unit (light blue box in Fig.~\ref{fig:architecture}) performs a GICP-based scan-to-scan and scan-to-submap matching operation to estimate the 6-DOF motion of the robot between consecutive lidar acquisitions. 

\ph{Notation}
We denote with $\mathcal{R}$ the coordinate system of the robot and with $\mathcal{W}$ the coordinate system of the world, that coincides with $\mathcal{R}$ at the start of the test. We therefore address the problem of determining the poses of $\mathcal{R}$ with respect to $\mathcal{W}$ by means of consecutive lidar acquisitions to incrementally reconstruct a trajectory and a map of the explored environment. We denote with $L_{k}$ the lidar scan collected at time $k$ and with $L_{k-1}$ the lidar scan collected at time $k-1$. All lidar scans are expressed in the robot frame. We denote with $\textbf{X}_{k} \in SE(3)$ the robot pose in $\mathcal{W}$ at time $k$ and with $\textbf{X}_{k-1}$ the robot pose in $\mathcal{W}$ at time $k-1$. We denote with $\textbf{T}^{k-1}_{k} = \textbf{X}^{-1}_{k-1}\textbf{X}_k$ the rigid body transformation between two consecutive robot poses, where $\textbf{T}^{k-1}_{k} \in SE(3)$ is the transform between $\textbf{X}_{k-1}$ and $\textbf{X}_{k}$. 

\vspace{0.7cm}

\subsubsection{Sensor integration module}
In robots with multi-modal sensing, if available, we use an initial transform estimate from a non-lidar source in the scan-to-scan matching stage to improve accuracy and reduce computation. 

\ph{Health monitoring}
Multiple sources of odometry (e.g VIO, KIO, WIO) and raw IMU measurements are first transformed into $\mathcal{R}$, and then fed into a health monitor which selects an output from a priority queue of inputs that are deemed healthy (see bottom left of Fig.~\ref{fig:architecture}). The system is designed to take in a variety of sources of health metrics to evaluate the health of input sources. For example, ongoing work is looking to integrate with the Heterogeneous Robust Odometry (HeRO) system~\cite{Santamaria-navarro2019} that employs custom health analysis (such as feature counts and observability analysis) on each odometry source, as well as rate and covariance checks. For our implementation presented below, we use a simple rate-check: if input messages are at a sufficient rate ($> 1$Hz), then the source is healthy. 

\ph{Priority queue}
The priority queue is intend to always select the highest accuracy source, based on previous testing for a given robotic system in similar environments. If the robot enters an area where the highest priority source is degraded, it is intended for this to be reflected in the health metric, that would trigger a transition to the next highest, healthy input. With this health-metric-driven dynamic switching, the priority queue is static. The priority queue for our legged robot is: VIO, KIO, IMU, no input, and for our wheeled robot is: VIO (if present), WIO, IMU, no input. 

We define the pose estimate (with respect to $\mathcal{W}$) of the highest priority source that is found to be healthy as $\textbf{Y}$. To reduce operations, we buffer only $\textbf{Y}$ and interpolate the buffered data at lidar timestamps, $t_{k-1}$, and $t_k$ to get $\textbf{Y}_{k-1}$ and $\textbf{Y}_{k}$: the pose of highest priority healthy source at times $t_{k-1}$ and $t_k$, respectively. We denote with $\textbf{E}^{k-1}_{k} = \textbf{Y}^{-1}_{k-1}\textbf{Y}_k$ the rigid body transformation of the sensor integration module output between $\textbf{Y}_{k-1}$ and $\textbf{Y}_{k}$ in the $[t_{k-1}, t_k]$ time interval where $\textbf{E}^{k-1}_{k} \in SE(3)$. Each odometry source provides a rotation and translation, whereas for the IMU we only use the rotation measurements. 

\vspace{0.7cm}

\subsubsection{Scan-to-scan}
In the scan-to-scan matching stage, GICP computes the optimal transformation $\hat{\textbf{T}}^{k-1}_{k}$ that minimizes the residual error $\mathcal{E}$ between corresponding points in $L_{k-1}$ and $L_{k}$. 
\begin{align}
\hat{\textbf{T}}^{k-1}_{k} = \arg\min_{\textbf{T}^{k-1}_{k}} \mathcal{E} (\textbf{T}^{k-1}_{k}L_{k}, L_{k-1}) \label{eq:scanscan}
\end{align}
When the sensor integration module is successful, we initialize the GICP with $\textbf{T}^{k-1}_{k} = \textbf{E}^{k-1}_{k}$. If all sensors fail, the GICP is initialized with identity rotation and zero translation and the system reverts to pure lidar odometry.

\vspace{0.7cm}

\subsubsection{Scan-to-submap}
The motion estimated in the scan-to-scan matching stage is further refined by a scan-to-submap matching step. Here $L_k$ is matched against a local submap $S_k$ which is taken from the local region of the global map $M_k$ given the current estimate of the robot pose in $\mathcal{W}$. 
\begin{align}
\tilde{\textbf{T}}^{k-1}_{k} = \arg\min_{\textbf{T}^{k-1}_{k}} \mathcal{E} (\textbf{T}^{k-1}_{k}L_{k}, S_{k})\label{eq:scanmap}
\end{align}
In this optimization, $\textbf{T}^{k-1}_{k}$ is initialized with the $\hat{\textbf{T}}^{k-1}_{k}$ result from Eqn.~\ref{eq:scanscan}. The global map is a point cloud stored in an octree format that is an accumulation of point clouds after every $t$~meters of translation, or $r$~degrees of rotation: for our results, we use $t = 1$, $r = 30^o$. We use an octree with a minimum resolution of 0.001m to store the map, which usually maintains all points in an easily searchable format. 

\ph{Output}
After scan-to-scan and scan-to-submap matching, the final estimated motion $\tilde{\textbf{T}}^{k-1}_{k}$ between consecutive lidar acquisitions is used to update the robot pose in $\mathcal{W}$: the generated odometry is therefore the integration of all computed incremental transforms. 

Both accuracy and computational speed are improved by the incremental estimation from input odometry to scan-to-scan and finally scan-to-map (shown, for example, in~\cite{LAMP}). A good initial estimate in both Eqn.~(\ref{eq:scanscan}) and Eqn.~(\ref{eq:scanmap}) reduces the chances of converging in a sub-optimal local minima, and reduces the number of iterations needed to converge, lowering computation time. 

\vspace{0.7cm}

\subsubsection{Notes on multi-threading}
The computational speed of the scan-to-scan and scan-to-submap matching has been greatly increased through the development of a multi-threaded GICP approach, modified from the PCL implementation~\cite{rusu20113d}. The multi-threading utilizes a user-specified number of cores for the normal computation stage in GICP, which represents over 70\% of the computation time in the process. For the evaluations performed in this paper, we use 4 threads unless otherwise stated. 

\subsection{Environment Adaptation: Flat Ground Assumption}
In human-made environments there are many areas with flat grounds, which if known prior, could be utilized to aid odometry algorithms. When detected or known, the flat ground assumption (FGA) can be activated to limit drift in Z and error in roll and pitch (lower-right blue box in Fig.~\ref{fig:architecture}). FGA operates on the output of both scan-to-scan and scan-to-submap alignment, by zeroing any Z movement, roll or pitch, all in a global, gravity aligned reference frame. 

\ph{FGA activation modalities}
The system provides two ways to detect a flat ground and activate FGA: context-based, and sensor-based. The first approach relies on prior knowledge of the environment that can be acquired by a human supervisor, for instance in single floor exploration of urban environments. For stair-climbing robots, the initiation of a stair mission can be used to deactivate FGA, and then reactivate it when the stair mission is complete through the input of a human operator (see~\cite{SpotPaper} for an example). In the second approach, an IMU monitor can be used to detect periods when the robot has near-zero roll and pitch over a sufficient time period to activate FGA, and then deactivate it when this condition is no longer met.

\subsection{Adaptation for Different Platforms}
The system includes adjustable components to adapt to heterogeneous robotics platforms with different computing power and sensors. These adaptations are primarily in: the number of lidars, the filtering, and number of threads for GICP and the measurements used for the initial transform estimate. Sec.~\ref{sec:real-time_results}, demonstrates the flexibility of LOCUS through application to two different robotic platforms.


\section{Field Experiments}\label{sec:field_demos}
In this section we present the experimental results obtained from tests in the Tunnel and Urban circuits of the SubT Challenge. We first use three datasets from a Clearpath Husky ground rover (Fig.~\ref{fig:cover}.a-b) to perform an ablation study on LOCUS, and compare it with state-of-the-art lidar odometry solutions. 
We then showcase results achieved during live operations in field tests across heterogeneous robotic platforms. See \url{https://youtu.be/5QQkkQ_YrbU} for visualization of the results.

\ph{Dataset description}
Each dataset comprises 3D lidar scans coming from 2 on-board VLP16 lidars (one flat, one pitched forward $30^o$), along with IMU (Vector Nav 100) and WIO measurements for a 60-minute run. Each dataset is selected to contain components that are challenging for lidar odometry. The Urban datasets (Alpha Course, Fig.~\ref{fig:cover}.f and Beta Course, Fig.~\ref{fig:cover}.e) are collected in a dismissed power plant located in Elma, WA that presents many challenges for robot perception such as long feature-poor corridors and large open spaces (the test area dimensions are 100x100x20 m). The Tunnel dataset (Safety Research Course, Fig.~\ref{fig:cover}.c) is recorded in the Bruceton Research Mine in Pittsburgh, PA that is characterized by self-similar and self-repetitive geometries (the test are dimension are 200x200x10 m). All datasets have substantial vibrations and large, sudden accelerations as is characteristic of a skid-steer wheeled robot traversing rough terrain and rubble. See Fig.~\ref{fig:cover} for sample images of the environments.

Lidar scans are recorded at 10Hz. WIO and IMU are recorded at 50Hz in the Urban datasets, while a higher-rate IMU recording (100Hz) is available for the Tunnel dataset. Both motion corrected and raw points are available for the Urban datasets, whereas the Tunnel dataset only has raw points available. 
We use LOCUS to do scan matching on the ground-truth map provided by DARPA to estimate the ground-truth reference of the robot trajectory. 

\subsection{Ablation Study}
\begin{figure}[hbt]
    \hspace{0.2em}
    \centering
    \includegraphics[trim={0cm 0cm 0cm 0cm},clip, width=0.9\columnwidth]{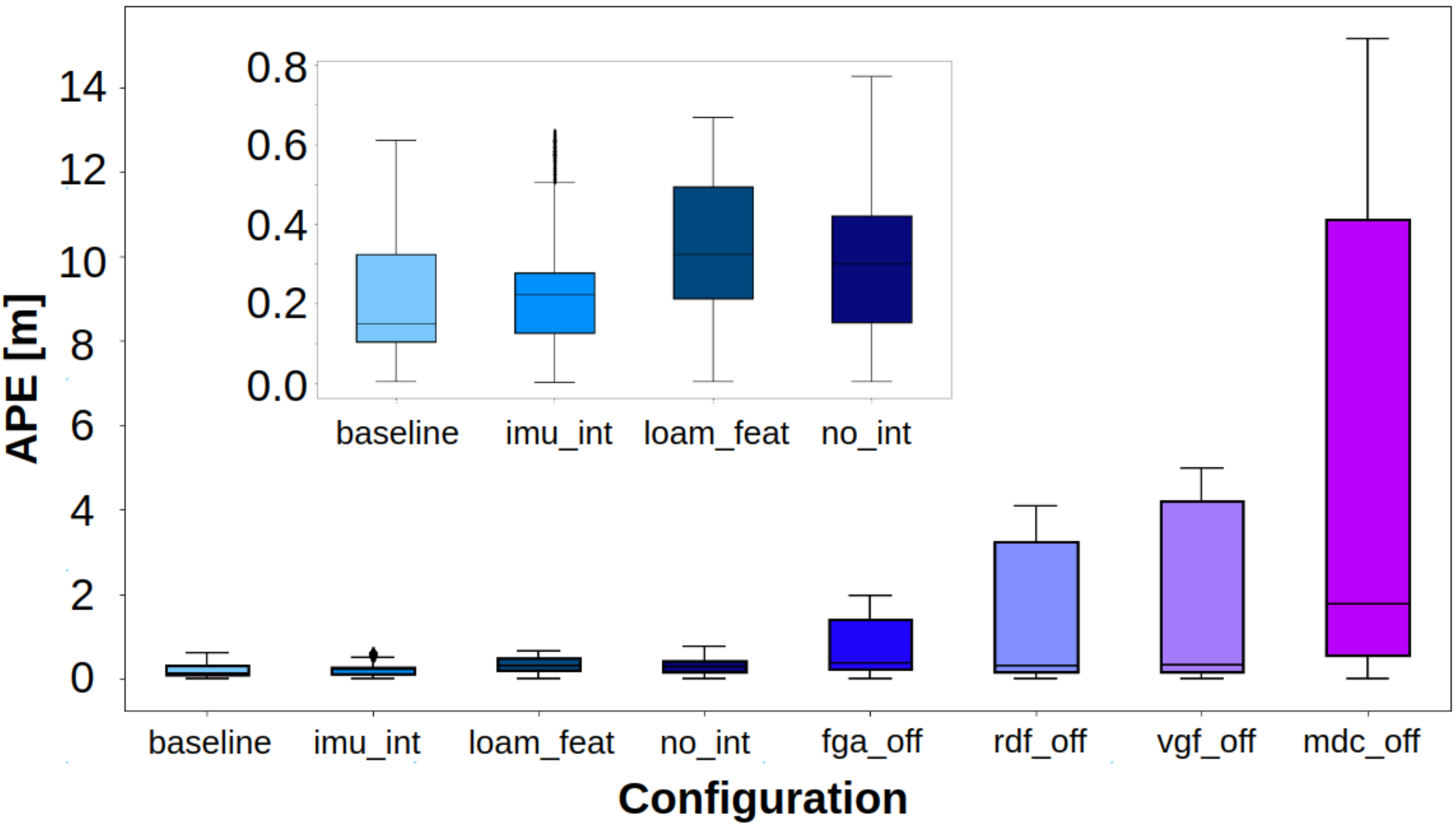}
    \caption{Evolution of the Absolute Position Error (APE) of the proposed method for different processing configurations in the Alpha course of the SubT Challenge. The inset gives more detail on the four best configurations. \textbf{baseline}: all features in Sec.~\ref{sec:locus}. \textbf{imu\_int}: no WIO integration, only IMU integration, \textbf{no\_int}: neither WIO or IMU integration, \textbf{loam\_feat}: using LOAM feature extraction instead of filtering, \textbf{fga\_off}: no FGA, \textbf{rdf\_off}: no random downsample filter, \textbf{vgf\_off}: no voxel-grid filter, \textbf{mdc\_off}: no MDC. }
    \label{fig:AblationStudy}
    \vspace{-0.4cm}
\end{figure}

To investigate the impact of each component of LOCUS on the overall pipeline accuracy, we evaluate the Absolute Position Error (APE) of the robot-trajectory in the Urban Alpha dataset\footnote{While these results are for one dataset, we observed similar trends from tests on the Urban Beta dataset. }
The results are summarized in Fig.~\ref{fig:AblationStudy}. 
The study confirms that the use of motion-corrected points is essential, and highlights the effectiveness of the filtering approaches that limit the data volume and reduce computational load. Feature-based filtering (e.g. LOAM-type features of edges and planes) can result in greater accuracy, yet with a higher CPU load (25\% more than the baseline configuration), leading to non-real-time performance on our system. Environmental knowledge of a flat ground is not essential, but can help to improve accuracy for exploration of human-made buildings. The loose sensor integration of WIO or IMU results in minor improvements, however, we use this approach as baseline to robustly operate in scenarios with high-rate motions or low lidar observability.

\begin{table}[]
\caption{Summary of State-of-the-Art, Open-Source Algorithms} 
\label{tab:algorithms}
\vspace{-0.2cm}
\centering
\begin{tabular}{|l|lllll|} \hline
Algorithm         & Align. & Opt.  & IMU & Odom. &MDC*        \\ \hline
LOCUS        & Dense         & GICP         & Loose    & Loose     & Yes                \\
BLAM\cite{BLAM}\tablefootnote{\url{github.com/erik-nelson/blam_slam}}         & Dense         & GICP         & None     & None         & No                 \\
ALOAM\cite{LOAM}\tablefootnote{\url{github.com/HKUST-Aerial-Robotics/A-LOAM}}        & Features      & Ceres       & None     & None    & No                        \\
FLOAM\cite{LOAM}\tablefootnote{\url{github.com/wh200720041/floam}}        & Features      & Ceres       & None     & None               & No            \\
Cartog.\cite{Cartographer}\tablefootnote{\url{github.com/cartographer-project/cartographer}} & Grid          & Ceres       & Tight    & Loose  & Yes                      \\
LIO-Map.\cite{LIO_Mapping}\tablefootnote{\url{github.com/hyye/lio-mapping}}  & Features      & Ceres       & Tight    & None     & Yes                \\
LIO-SAM\cite{liosam2020shan}\tablefootnote{\url{github.com/TixiaoShan/LIO-SAM}}      & Features      & GTSAM      & Tight    & None     & Yes    \\ \hline

\multicolumn{6}{l}{* See section IIIB for more details.} 

\end{tabular}
\vspace{-0.5cm}
\end{table}

\subsection{Evaluation Against the State-of-the-Art}
We compare the proposed algorithm against a variety of the state-of-the-art open-source lidar odometry systems, selected to cover the range of aligment methods and sensor integration methods, as summarized in Table~\ref{tab:algorithms}. 
While we would like to compare against systems integrating with visual odometry (e.g.~\cite{VLOAM},~\cite{VIL_SLAM}), these algorithms do not have open source implementations to readily test. 
FLOAM and ALOAM are two modern implementations of LOAM aimed to simplify the code structure and increase computational speed, respectively. Cartographer is an LIO algorithm distinguished by its use of grid-based matching. LIO-Mapping is a more recent LIO algorithm that combines the IMU pre-integration approach from VINS-Mono~\cite{qin2018vins} with LOAM-type feature alignment. LIO-SAM is yet more recent, and builds from LeGo-LOAM~\cite{LeGO-LOAM}, adding in IMU pre-integration in a smoothing-and-mapping approach.

\ph{Comparison criteria}
We aim to compare the lidar odometry systems holistically, hence we use three criteria: i) Accuracy, ii) Robustness and iii) Efficiency. 

\ph{Comparison setup} 
Each algorithm is setup for the best performance, yet with minimal parameter tuning, only input adjustment (number of lidars, motion corrected points). WIO is the odometry input, if used. Loop closures are disabled to focus on the lidar odometry performance. Variations in the input for each algorithm are summarized below. 

LOCUS, BLAM, and ALOAM each use two lidars in all datasets, with motion-corrected points in the Urban datasets. FLOAM only succeeded with one lidar in the Urban Alpha dataset and otherwise uses two lidars, in each case (with motion correction in Urban datasets). Cartographer can do internal motion correction, but only with the input of points as individual UDP packets from a single lidar. We elected to instead feed pre-corrected scans from two lidars to Cartographer. LIO-Mapping and LIO-SAM are both set up to do internal motion-correction on the point clouds, leveraging the integrated IMU data, hence uncorrected scans are used as inputs. For LIO-Mapping, 2 lidars were used, except for Urban Alpha, where only a 1 lidar test was successful. 

LIO-SAM was only able to run with a single lidar input, as there are internal assumptions of a structured point cloud of rings for motion correction and feature extraction. We were not able to get LIO-SAM working on the Urban datasets, likely due to the IMU rate, at 50Hz being lower than the recommended 200Hz for LIO-SAM. 

\subsubsection{Accuracy Evaluation}
We evaluate the accuracy with two metrics: position error and map error. 

\begin{figure*}
    \hspace{0.2em}
    \centering
    \includegraphics[trim={0cm 0cm 0cm 0cm},clip, width=2\columnwidth]{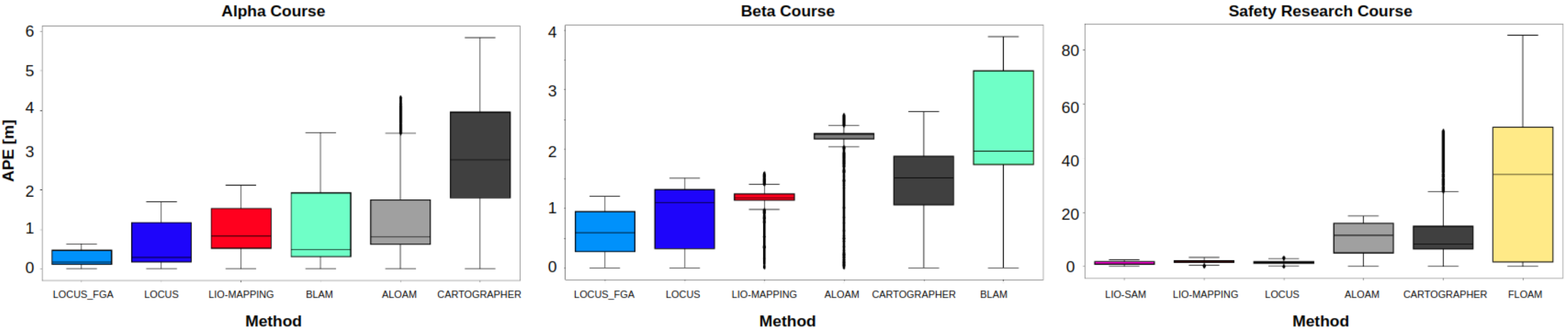}
    \caption{Boxplot visualization of the Absolute Position Error (APE) computed for the different methods on the test datasets. For clarity, only the best six algorithms in each dataset are shown.} 
    \label{fig:OdometryAccuracyBoxplots}
    \vspace{-0.0cm}
\end{figure*}

\ph{Position error}
To evaluate the localization accuracy, we use evo~\cite{grupp2017evo} to compute the absolute position error (APE) of the trajectories estimated by the different methods against the ground-truth reference. We report in Fig.~\ref{fig:OdometryAccuracyBoxplots} a boxplot visualization of the APE results for the Urban and Tunnel datasets and summarize these results in Table~\ref{tab:accuracy_table}.

The results show that LOCUS is equal to or better than the state-of-the-art in all datasets. FGA does help to improve LOCUS performance, but is not essential for LOCUS to perform well. LIO-Mapping has similarly low error, as expected with tight integration of IMU and lidar, yet with a large delay from a mean processing time of 1s per scan. The larger optimization being performed with pre-integration, scan alignment and extrinsic estimation all together likely leads to the longer computation times. 

LIO-SAM performs the best in the Tunnel dataset, and with feasible computational speeds, yet the strict requirements on appropriate input data limit the range of platforms and datasets it can be applied to. Cartographer and ALOAM perform relatively well in Urban, showing the effectiveness of edge and plane features as well as grid methods in human-made environments. 

For dense alignment methods, BLAM perfoms adequately in the Urban datasets, but poorly in the Safety Research dataset, suggesting the WIO integration employed by LOCUS is an important component in the self-similar and low lidar observability conditions seen in that dataset. 

\ph{Map error}
As second quantitative evaluation, we compare the maps obtained with each algorithm to the DARPA provided ground-truth map to compute the overall cloud-to-cloud error. To account for potential calibration misalignments, we run Iterative Closest Point (ICP) between the reconstructed map and ground-truth map before performing error analysis.
We report in Table~\ref{tab:accuracy_table}, a numerical summary of the RMSE values of the map error (ME) computed for each algorithm on each relevant dataset. The results show similar trends to the position error, with some negligible differences in the order of the algorithms 

\begin{table*}[]
\caption{Summary of Accuracy Analysis results on Alpha, Beta and Safety Research datasets} \label{tab:accuracy_table}
\vspace{-0.3cm}
\begin{tabular}{|l|l|l|l|c|l|l|l|c|l|l|l|c|c|c|}
\hline
       \textbf{Algorithm}              & \multicolumn{4}{c|}{\textbf{Alpha Course}}                                                                 & \multicolumn{4}{c|}{\textbf{Beta Course}}                                                                  & \multicolumn{4}{c|}{\textbf{Safety Research Course}}                                                     & \multicolumn{2}{c|}{\textbf{CPU}$^\beta$}    \\ \hline
                      & \multicolumn{3}{c|}{APE {[}m{]}}                                         & \multicolumn{1}{l|}{ME {[}m{]}} & \multicolumn{3}{c|}{APE {[}m{]}}                                         & \multicolumn{1}{l|}{ME {[}m{]}} & \multicolumn{3}{c|}{APE {[}m{]}}                                         & \multicolumn{1}{l|}{ME {[}m{]}} & \multicolumn{2}{c|}{Num. of Cores}  \\ \hline
                      & max                    & mean                   & std                    & \multicolumn{1}{l|}{RMSE}       & max                    & mean                   & std                    & \multicolumn{1}{l|}{RMSE}       & max                    & mean                   & std                    & \multicolumn{1}{l|}{RMSE}   & max & mean  \\ \hline
\textbf{LOCUS}        & 1.69                 & 0.62                 & 0.57                 & 0.29                          & 1.51        & 0.88       & 0.51       & 0.69                               & 3.39        & 1.67        & 0.76        & 0.63           & 3.39 & 2.72                       \\ \hline
\textbf{LOCUS\_FGA}   & \textbf{0.63}                 & \textbf{0.26}                 & \textbf{0.18}                & \textbf{0.28}                          & \textbf{1.20}       & \textbf{0.58}        & 0.39       & 0.48                               & \multicolumn{1}{c|}{-} & \multicolumn{1}{c|}{-} & \multicolumn{1}{c|}{-} & -              & 3.39 & 2.72                   \\ \hline
\textbf{BLAM}         & 3.44                 & 1.01                 & 0.94                 & 0.43                               & 3.89                 & 2.27                 & 0.89                 & 1.27                               & 171.34                 & 35.45                 & 51.91                 & 5.37                          & \textbf{1.14} & 0.93         \\ \hline
\textbf{ALOAM}        & 4.33                 & 1.38                 & 1.19                 & 0.60                          & 2.58                 & 2.11                 & 0.44                 & 0.99                          & 18.61                 & 10.01                 & 6.01                 & 6.11                        & 1.65 & 1.41      \\ \hline
\textbf{FLOAM}        & 29.49                 & 9.19                 & 8.96                 & 1.73*                          & 40.64                 & 3.94                 & 8.42                 & 3.73*                          & 85.31                 & 32.49                 & 25.73                 & 20.16                         & 1.76 & 1.44    \\ \hline
\textbf{Cartographer} & 5.84                 & 2.91                 & 1.60                 & 1.05                          & 2.64                 & 1.37                 & 0.67                 & \textbf{0.31}                          & 50.05                 & 14.31                 & 13.45                 & 14.25                       & 1.75 & \textbf{0.88}      \\ \hline
\textbf{LIO-Mapping}  & 2.12                 & 0.99                 & 0.51                 & 0.45                          & 1.60                 & 1.18                 & \textbf{0.22}                 & 0.61                          & 3.31                 & 1.99                 & \textbf{0.55}                 & 0.76                      & 1.80 & 1.53         \\ \hline
\textbf{LIO-SAM}      & \multicolumn{1}{c|}{-} & \multicolumn{1}{c|}{-} & \multicolumn{1}{c|}{-} & -                               & \multicolumn{1}{c|}{-} & \multicolumn{1}{c|}{-} & \multicolumn{1}{c|}{-} & -                               & \textbf{2.45}                 & \textbf{1.26}                 & 0.58                 & \textbf{0.52}                        & 2.75$^+$ & 2.00$^+$          \\ \hline
\multicolumn{15}{l}{* failure leads to a low map error. $^\beta$ CPU loads are computed from the Urban Beta dataset, and Tunnel dataset for LIO-SAM ($^+$)}
\end{tabular}
\end{table*}

\subsubsection{Robustness Evaluation}
\begin{figure}
    \hspace{0.2em}
    \centering
    \includegraphics[angle=0,trim={0cm 0cm 0cm 0cm},clip, width=0.9\columnwidth]{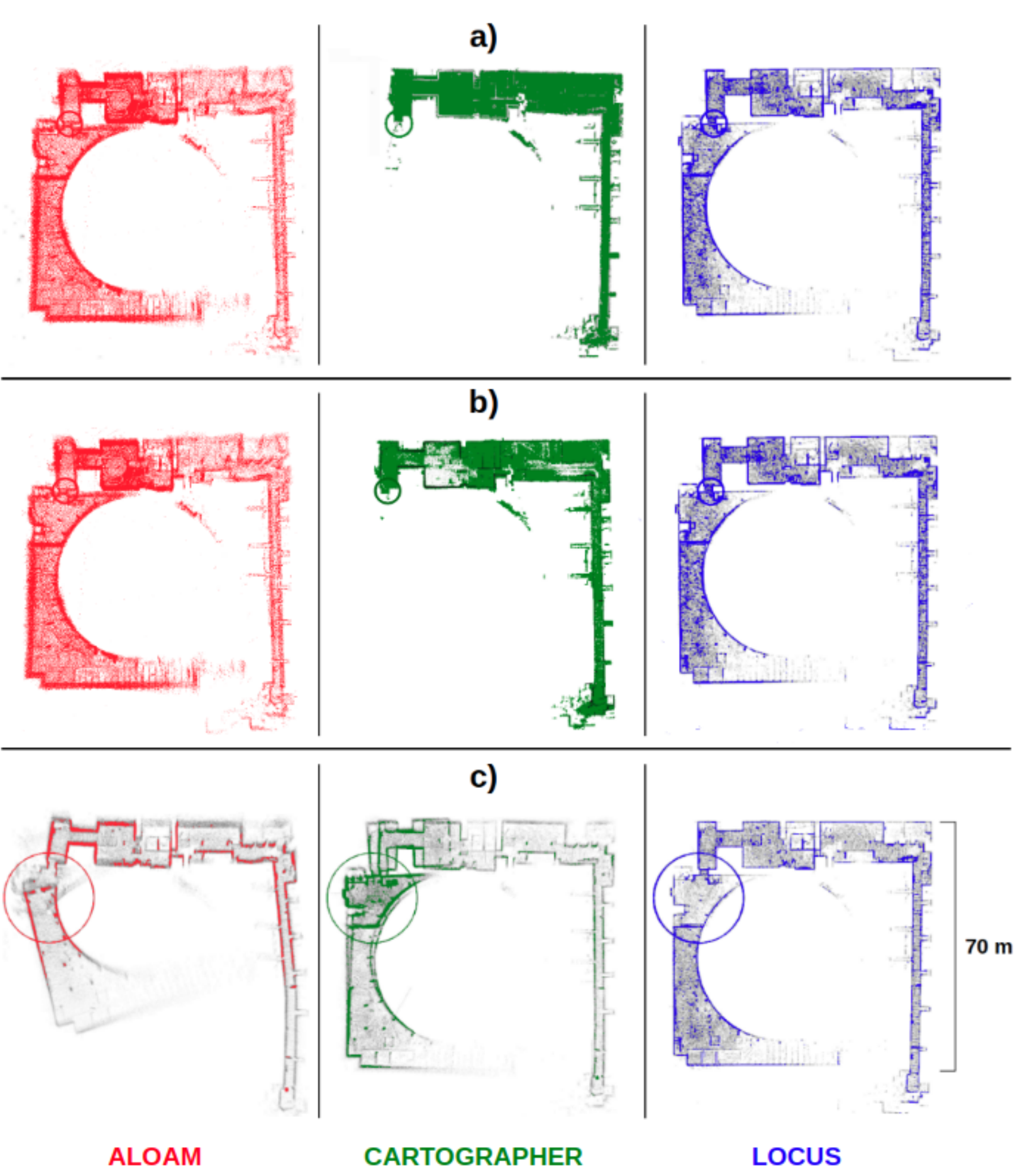}
    \caption{Robustness test in Beta course: a) results on WIO/IMU failure, b) results on WIO failure, c) results on Lidar failure. The failure locations are circled in all cases.}
    \label{fig:Robustness}
    \vspace{-0.7cm}
\end{figure}

The previous section highlighted the robustness to low lidar observability, substantial vibrations, large accelerations and self-similar environments through the accuracy results on the datasets. In this section we focus on another aspect of robustness: the ability to handle a sudden failure of an input source. Specifically, we test the following scenarios: i) failure of WIO/IMU, ii) failure of WIO, iii) failure of lidar. Each of these failure scenarios have been experienced in real field tests in preparation for the SubT Challenge. We artificially create these failures in our datasets to have a controlled way of isolating the source of the failure, and the resulting impact on the algorithms. The results are summarized in Table~\ref{tab:robustness_table}, with example maps resulting from different failure modes shown in Fig~\ref{fig:Robustness}.

\begin{table}[]
\caption{Summary of Robustness Test Results} 
\label{tab:robustness_table}
\vspace{-0.3cm}
\begin{tabular}{|l|lll|} \hline
             & \multicolumn{3}{c|}{Robustness Test Result}                \\
Algorithm    & a) WIO/IMU Fail        & b) WIO Fail & c) Lidar Drop \\ \hline
LOCUS        & OK                     & OK          & OK            \\
BLAM         & NA             & NA  & Errors        \\
ALOAM        & NA             & NA  & Errors        \\
FLOAM        & NA             & NA  & Errors        \\
Cartographer & Stops                  & Stops       & Errors        \\
LIO-Mapping  & Stops                  & NA  & Errors        \\
LIO-SAM      & Stops                  & NA  & Errors       \\ \hline
\multicolumn{4}{p{8cm}}{\tiny{\textbf{OK:} negligible degradation in accuracy. \textbf{NA:} Not Applicable - the algorithm does not use that sensor source. \textbf{Errors:} substantial errors in accuracy. \textbf{Stops:} no more odometry output after failure.}}
\end{tabular}
\vspace{-0.7cm}
\end{table}

\ph{WIO/IMU failure}
We simulate sensor failure after 1200s, by shutting down WIO and IMU streams for the rest of the run. This failure only affects those algorithms that use IMU, where the algorithms cease to run, either relying on a synced callback with WIO and IMU (e.g. Cartographer) or relying on pre-integrated IMU to provide odometry updates between scans as well as initial scan to map alignment estimates (LIO-Mapping, LIO-SAM). In contrast, LOCUS processes the input data separately, and hence can automatically switch from lidar odometry with WIO integration, to lidar odometry with IMU integration, to pure lidar odometry, demonstrating robust handling of sensor failures in a cascaded fashion. This behavior is highly desirable to accommodate the unforeseen challenges posed by rough terrains in real-world applications where hardware failures are likely to happen, or sensors sources can become unreliable (e.g. dark areas with no visual texture for VIO). 

\ph{WIO failure}
We simulate a loss of WIO after 1200s. Cartographer and LOCUS are the only algorithms affected, with the same result as the WIO/IMU case.

\ph{Lidar failure}
We stress the systems by subtracting the most fundamental data source: lidar. More specifically, we simulate a 10s gap in lidar data while the robot is in motion.

There are three responses to this failure. The first response is that the algorithm stops running until the lidar returns, resulting in large map errors (BLAM, ALOAM, FLOAM, and Cartographer due to the synced callback). The second response is that the algorithm runs purely on IMU integration, leading to an accumulation of drift before the lidar returns (LIO-Mapping, LIO-SAM). The final response is only seen by LOCUS, where the loose coupling allows WIO to accumulated over the 10s of no lidar data to produce an accurate initial transform when the lidar returns.

\subsubsection{Efficiency Evaluation}
\begin{figure}
    \centering
    \includegraphics[trim={0cm 1.2cm 0cm 0cm},clip, width=0.75\columnwidth]{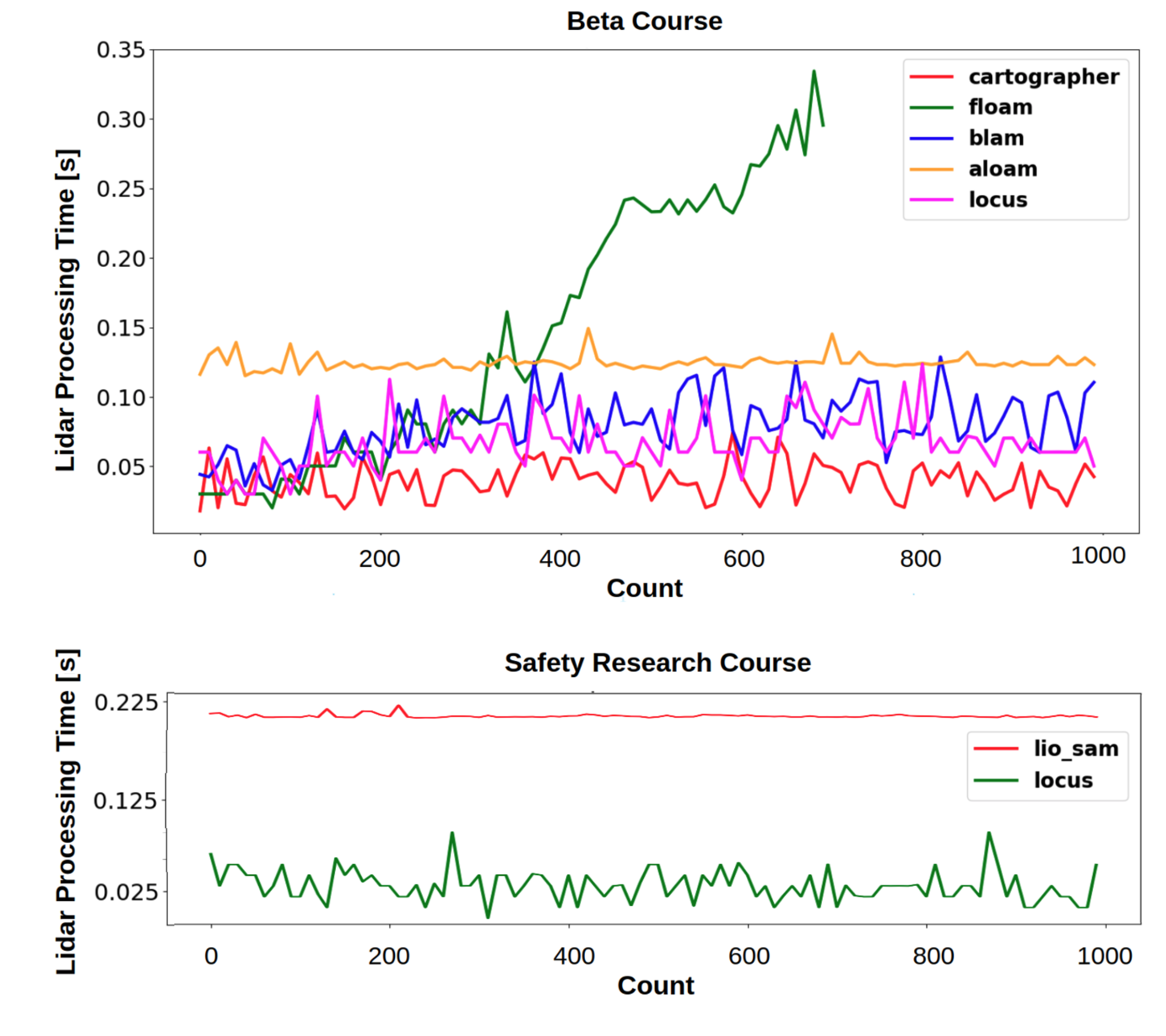}
    \caption{Comparison of lidar processing time across the different lidar odometry algorithms. The times are the duration for processing a single scan. Top - Urban Beta dataset, Bottom - Tunnel Safety Research dataset. A processing time of 0.1~s indicates realtime performance (10 Hz scans).}
    \label{fig:Time}
    \vspace{-0.65cm}
\end{figure}

We profile the time needed from the different algorithms to process a single lidar scan when running the algorithms on an Intel Hades Canyon NUC8i7HVKVA (4x1.9 GHz, 32 GB RAM) running Ubuntu 18.04 LTS. Fig.~\ref{fig:Time} shows the resulting times per scan with scans at 10Hz (LIO-Mapping is omitted as the processing time, 1s per scan, is too large). Additionally, Table~\ref{tab:accuracy_table} shows the CPU loads for each algorithm. All values are from the Urban Beta dataset, except for LIO-SAM, which is on the Tunnel Safety Research dataset. 

LOCUS, ALOAM and BLAM can all maintain real-time processing, whereas ALOAM and LIO-SAM can only stay real-time with a lower rate of lidar scans. LIO-SAM can use the IMU pre-integration to cope with a lower IMU rate, and by using features, ALOAM can also handle a lower rate for certain datasets. Both FLOAM and LIO-Mapping do not appear to be feasible for real-time operation. Cartographer has both the quickest processing time and the lowest CPU load, yet the accuracy is not as strong as the other algorithms. LOCUS produces the best accuracy, with real-time performance, yet requires the largest CPU load. 

\subsection{Real-Time Operation Across Different Platforms} \label{sec:real-time_results}
In this section, we demonstrate real-time field operation of LOCUS on different robotic platforms during the Urban Circuit of the SubT Challenge and provide statistics from logged online operation. Results come from the four competition runs, two in Alpha course and two in Beta course. 

\vspace{0.7cm}

\subsubsection{Hardware and Tuning}
During the competition, we deployed LOCUS on two very different robots: i) the Husky from the datasets used above (see Fig.~\ref{fig:cover}.b), and  ii) Spot from Boston Dynamics (see Fig.~\ref{fig:cover}.d). 

\ph{Husky}

In addition to the sensors described in Sec.~\ref{sec:field_demos}, Husky carries an AMD RYZEN 9 3900X 12-Core 3.8 GHz for computation.

\ph{Spot}
A legged robot that is equipped with 1 VLP16 and an Intel NUC7i7DN 4-Core 1.9 GHz for computation. Both VIO and KIO are available from the Boston Dynamics API, and can be used for integration into LOCUS. We choose VIO as it was shown to be more accurate than KIO in our tests. 

\ph{Adaptation}
The paramaters of LOCUS are tuned to achieve accurate and robust real-time operation on both platforms while accounting for differences in computational capabilities and hardware configurations. Table~\ref{tab:settings} summarizes the configurations used during the competition. 

\begin{table}[h]
\caption{Summary of LOCUS settings on different robots} 
\label{tab:settings}
\vspace{-0.3cm}
\centering
\begin{tabular}{|c|c|c|}
\hline Parameter & \textbf{Husky} & \textbf{Spot} \\ \hline
Number of lidars                  & 2                               & 1                              \\ \hline
Voxel Grid Filter leaf size (m)    & 0.1                             & None                           \\ \hline
GICP iterations in scan-to-submap & 20                              & 25                             \\ \hline
GICP number of cores              & 4                               & 1                              \\ \hline
Sensor Integration                & WIO                             & VIO       \\ \hline 
\end{tabular}
\end{table}

\subsubsection{Performance}
We report in Table~\ref{tab:dropped scans} the average value of the number of lidar scans dropped per second by each robot in each course of the competition during real-time operation. Point clouds are subscribed to at 10hz and we do not buffer any lidar scans, to minimize the delay of the computed odometry. Therefore, the number of dropped scans per second represent how frequently the lidar processing time exceeded 0.1s. Spot drops 2 scans a second, due to the less powerful computer onboard. However, Spot has a more accurate additional odometry source than Husky, with VIO. The reliable initial transform from VIO enables Spot to process fewer scans per second, and still perform well. 

\begin{table}[h]
\caption{Dropped lidar scans from real-time on-robot tests} 
\label{tab:dropped scans}
\vspace{-0.3cm}
\centering
\setlength{\tabcolsep}{8.3pt}
\begin{tabular}{|c|c|c|c|c||c|}
\hline
                   & \multicolumn{5}{c|}{\textbf{Number of dropped scans / s}}                                    \\ \cline{2-6} 
\multirow{-2}{*}{Robot} & Alpha 1 & Alpha 2 & Beta 1 & Beta 2 & {Average} \\ \hline
    \textbf{Husky}     & 0       & 0       & 0      & 0      & {0}       \\ \hline
\textbf{Spot}      & 2.082       & 2.205       & 1.833      & 2.016  & 2.034   \\ \hline
\end{tabular}
\end{table}

\begin{figure}
    \centering
    \includegraphics[trim={0cm 0cm 0cm 0cm},clip, width=0.75\columnwidth]{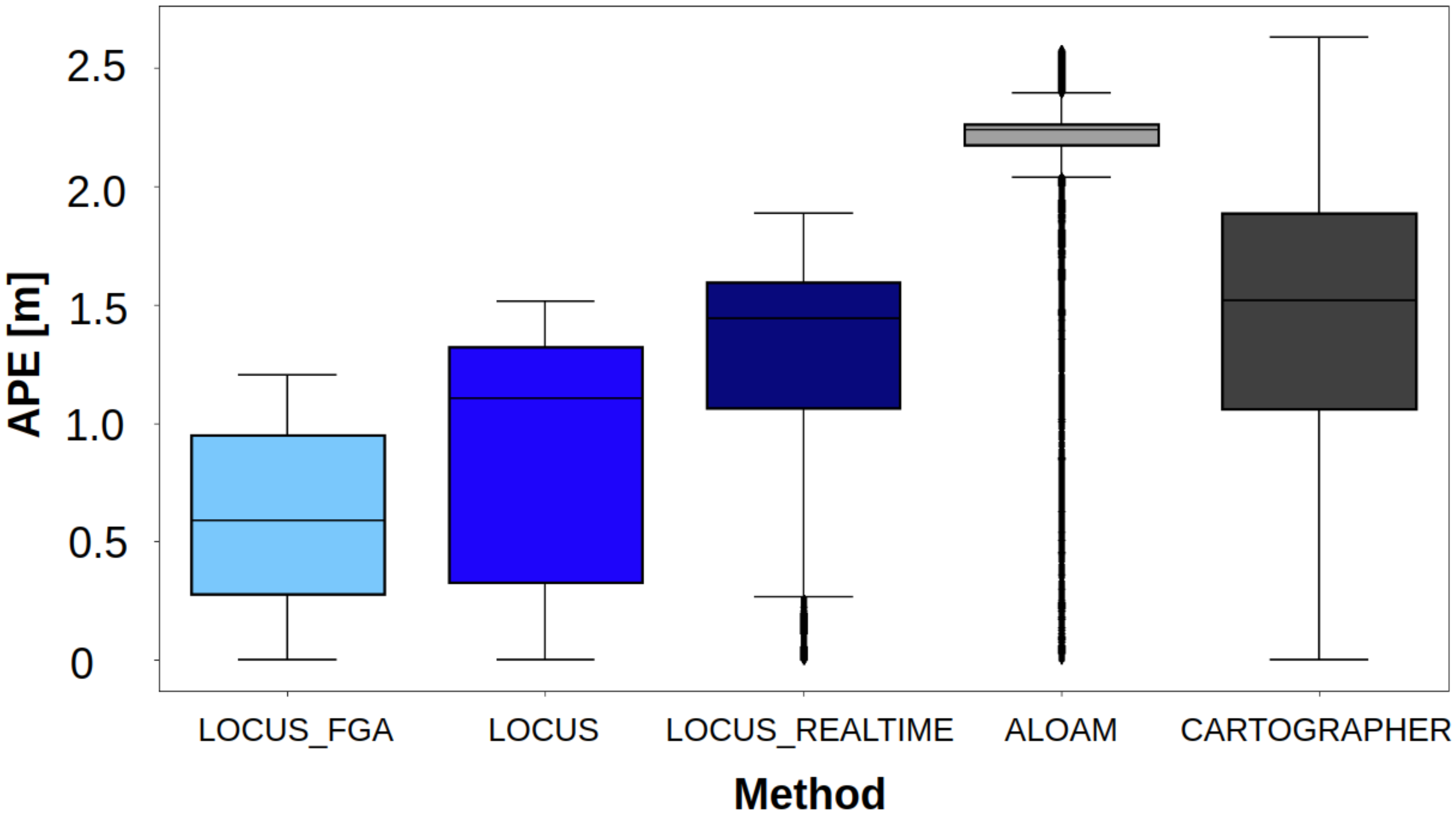}
    \caption{Absolute Position Error (APE) of the trajectories estimated by the different methods against ground-truth in Beta course, including the performance when running live in the SubT Challenge (LOCUS\_REALTIME).}
    \label{fig:AllBoxplotResults}
    \vspace{-0.4cm}
\end{figure}

\ph{Real-time accuracy profiling}
LOCUS performed accurately for both Husky and Spot in the competition, as evident in the overall team's performance, winning first place\footnote{The LOCUS output was integrated with a robust odometry aggregator~\cite{Santamaria-navarro2019}, and then fed to a back-end SLAM algorithm~\cite{LAMP}}. Fig.~\ref{fig:AllBoxplotResults}, shows the live performance of LOCUS with FGA on the Husky in Beta 2, with a slightly larger error than the post-processed results (on a different computer), yet still highly competitive. 

For Spot, LOCUS was run live in a multi-level exploration in the Urban Alpha 2 course. In this run, the mean APE was $0.586$ m, and the maximum APE was $2.599$ m, which was sufficiently small for scoring in the competition.

\subsection{Discussion}
Overall, fusion of additional sensing modalities is crucial to enable accurate operation in such extreme explorations: by relying on a loosely-coupled mechanism, LOCUS is robust to potential failures of sensors, and can achieve improved performance with respect to tightly-coupled approaches in settings where the extrinsic sensors calibration is not ideal. Additionally, by not making assumptions on the environment type, LOCUS can use a larger number of points with respect to feature-based methods during scan registration, and process a greater amount of information at reasonable computational cost by taking advantage of the priors informed by the additional sensing modalities to seed the GICP.


\section{Conclusions}
Achieving accurate lidar odometry in perceptually-challenging conditions can be difficult due to the lack of reliable perceptual features, presence of noisy sensor measurements, and high-rate motions. While integrating additional sensing modalities can help address these challenges, potential sensor failures can have dramatic impacts on the mission outcome if not robustly handled. In this paper, we present a lidar odometry system to enable accurate and resilient ego-motion estimation in challenging real-world scenarios. The proposed system, LOCUS, provides an accurate multi-stage scan matching unit equipped with an health-aware sensor integration module for seamless loose integration of additional sensing modalities. The proposed architecture is adaptable to heterogeneous robotic platforms and is optimized for real-time operation. 

We compare LOCUS against state-of-the-art open-source algorithms and demonstrate top-class accuracy in perceptually challenging real-world datasets, top-class computation time and superior robustness to sensor failures, yet with greater CPU load.
Finally, we demonstrate field-proven real-time operation of LOCUS on two different robots involved in fully autonomous exploration of Satsop power plant during the Urban Circuit of the DARPA Subterranean Challenge, where the proposed system was a key component of CoSTAR team's solution that achieved first place.


\bibliographystyle{IEEEtran}
\bibliography{IEEEexample}

\end{document}